\documentclass{article}

 \usepackage[preprint,nonatbib]{neurips_2021}

\usepackage[utf8]{inputenc} 
\usepackage[T1]{fontenc}    
\usepackage{hyperref}       
\usepackage{url}            
\usepackage{booktabs}       
\usepackage{amsfonts}       
\usepackage{nicefrac}       
\usepackage{microtype}      
\usepackage{xcolor}         

\usepackage{wrapfig}
\usepackage{enumitem}
\usepackage[style=numeric, url=false, doi=false, isbn=false, eprint=false]{biblatex}
\usepackage{amsmath}
\usepackage{amsthm}
\usepackage[ruled]{algorithm2e}
\usepackage{graphicx}
\usepackage{caption}
\usepackage[font=small]{subcaption}

\newcommand{\BR}{\mathbb{R}}
\newtheorem{theorem}{Theorem}[section]

\newtheorem{lemma}[theorem]{Lemma}
\theoremstyle{definition}
\newtheorem{definition}{Definition}[section]
\newtheorem*{remark}{Remark}
\DeclareMathOperator{\rank}{rank}
\DeclareMathOperator{\I}{I}

\title{Understanding Training-Data Leakage from Gradients in Neural Networks for Image Classification}

\author{Cangxiong Chen \\
  Department of Computer Science\\
  University of Bath \\
  \texttt{cc2458@bath.ac.uk} \\
  \And
  Neill D.\,F.~Campbell \\
  Department of Computer Science\\
  University of Bath \\
  \texttt{nc537@bath.ac.uk}
}

\begin{document}

\maketitle
\vspace{-6mm}
\section{Introduction}
In federated learning \cite{pmlr-v54-mcmahan17a} of deep learning models for supervised tasks such as image classification and segmentation, gradients from each participant are shared either with another participant or are aggregated at a central server. In many applications of federated learning, the privacy of the training data will need to be protected and we want to obtain guarantees that a malicious participant will not be able to recover fully the training data from other participants, with shared gradients and knowledge of the model architecture. The guarantee will be indispensable in removing the barriers for applying federated learning in tasks such as image segmentations in film post-production where the training data are usually under strict IP protections. In this scenario, it is the training data that needs to be protected, rather than the information we can infer about them. 

In order to develop protection mechanisms, an appropriate understanding of the source of leakage of the training data is needed. For this work, we are concerned with the following question: for a deep learning model performing image classifications, what determines the success of reconstructing the training data given its label, its gradients from training, and the model architecture? We will focus on the case when we aim to reconstruct a single target image with an untrained model. Although our work was inspired by R-GAP \cite{zhu2021rgap}, our method COPA (combined optimisation attack) provides a more general theoretical framework to training-data reconstructions, particularly for convolutional layers. Compared with DLG \cite{dlg2019}, COPA provides more insight to the mechanism of training-data leakage through a more informative formulation of the objective function, making it clearer the source of constraints. 
\section{The COPA method}
\paragraph{Overall assumptions}
We only consider the problem of reconstructing a single training image in this work. The target image-classification model is assumed to consist of consecutive convolutional layers with the last layer being fully-connected. We assume the ground-truth label of the target image is given. This does not lose generality when the activation is nonnegative because otherwise we can reconstruct the true label from the last layer via observing the signs of the gradients of the weights according to \cite{zhao2020idlg}. The sign of the gradients of the weights corresponding to the true class will be different from those corresponding to other classes.  We assume that the activation function for each layer is piecewise invertible and piecewise differentiable. For simplicity we do not include pooling in the design of the network and dimensionality reduction in the feature space is mainly achieved by taking strides to be greater than one. Notice that CNN without pooling was also considered in \cite{springenberg2015striving}.  

\paragraph{Notation}
We introduce notation to be used throughout the paper. Superscript with parenthesis indicates the index of a layer in the network.
\begin{enumerate}[itemsep=-1mm]
    \item[$W^{(i)}$:] weight from layer $i$ of size $m$ by $n$, where $0 \leq i \leq d$ and $d$ is the total number of layers. For simplicity of notation, we omit $i$ in $m$ and $n$ when possible, but readers should be aware that the size of the weight need not be the same for different layers. For a convolutional layer, it denotes the circulant representation of the kernel following \cite[chapter 4.8.2]{golub2013matrix}.
    \item[$J$:] composition of softmax function with negative log likelihood loss function.
    \item[$\nabla _{W^{(i)}} J$:] gradient of the cost function $J$ w.r.t. $W^{(i)}$.
    \item[$Z^{(i)}$:] the linear output of the layer $i$ before activation given by $W^{(i)} X^{(i)} + B^{(i)}$ with input $X^{(i)}$ and bias $B^{(i)}$. This also expresses the convolutional operation following the circulant form of $W^{(i)}$.
    \item[$A^{(i)}(\cdot)$:] activation function after linearity in vector form. We use $a^{(i)}$ to denote its component. 
    \item[$|.|$:] when applied to a matrix, the absolute value sign $|.|$ denotes the number of elements.
    \item[$w,x,z,b$:] we use small letters with subscripts to denote the component at specified indices of the corresponding matrix in capital letters.
\end{enumerate}
Our method COPA consists of reconstructions of input for a fully connected layer and a convolutional layer. First we formulate a linear system using weights and gradients.
\paragraph{Weight and gradient constraints:}
At a given layer $i$, the forward and backward propagations give rise to the following equations:
\begin{subequations}
\begin{align}
        & W^{(i)} X^{(i)} + B^{(i)} = Z^{(i)} \label{weightconstraint},  \\
        & \nabla _{Z^{(i)}} J \cdot X^{(i)}  = \nabla _{W^{(i)}} J \label{gradconstraint}. 
\end{align}
\end{subequations}
We note that this represents both the fully-connected and the convolutional cases, using the circulant representation for the weight $W^{(i)}$ and the gradient $\nabla _{Z^{(i)}} J$. Both $Z^{(i)}$ and $\nabla _{W^{(i)}} J$ are written as vectors. From the reconstruction point of view, we treat $X^{(i)}$ as the unknown and regard the above equations as weight and gradient constraints imposed on the unknown. The term $Z^{(i)}$ is computed from inverting the reconstruction from the subsequent layer $X^{(i+1)}$:
\begin{equation}
    Z^{(i)} = (A^{(i)})^{-1}(X^{(i+1)}).
\end{equation}
The term $\nabla _{Z^{(i)}} J$ can be computed by using the following relations deduced from backpropagation:
\begin{subequations}\label{gradrelations}
\begin{align}
    & \nabla _{Z^{(i)}}J = \nabla_{X^{(i+1)}}J \cdot \nabla_{Z^{(i)}} A^{(i)} \label{gradrelationsZ}, \\
    & \nabla_{X^{(i)}} J = \nabla_{X^{(i+1)}} J \cdot \nabla_{Z^{(i)}} A^{(i)} \cdot W^{(i)} \label{gradrelationsX}.
\end{align}
\end{subequations}
It is worth noticing that the circulant representation of the gradient $\nabla _{Z^{(i)}} J$ is determined by the circulant form of the weight $W^{(i)}$ from backpropagating through the weight constraint \eqref{weightconstraint}.

\paragraph{Fully connected layer}
For a fully connected layer, the input can be solved uniquely in closed form. This is because there is no weight sharing in a fully-connected layer. So the circulant form of $\nabla _{Z^{(i)}} J$ consists of blocks of diagonal matrices with the same element along the diagonal in each block, which allows \eqref{gradconstraint} to be solved exactly. The solution and its proof can be found in Appendix~\ref{appendReconFullyConnectedLayer}.

\paragraph{Convolutional layer}
For a convolutional layer, we can no longer uniquely determine the input in general because of weight sharing. We proceed by first combining the weight and gradient constraints \eqref{weightconstraint} 
\eqref{gradconstraint} into a quadratic function. More precisely, write $U^{(i)}, V^{(i)}$ to denote the following block matrices
\begin{equation}\label{DefUandV}
    U^{(i)} := \begin{bmatrix}
                 W^{(i)} \\
                 \nabla _{Z^{(i)}} J
                \end{bmatrix},
    V^{(i)} := \begin{bmatrix}
                (A^{(i)})^{-1}(X^{(i+1)}) \\
                \nabla _{W^{(i)}} J
                \end{bmatrix}.
\end{equation}
Here both the term $(A^{(i)})^{-1}(X^{(i+1)})$ and the term $\nabla _{W^{(i)}} J$ are written as vectors. The term $\nabla _{W^{(i)}} J$ has the same dimension as $|W^{(i)}|$, i.e.~the number of elements of the weight in its non-circulant form as a 4-dimensional array. Notice that we can absorb the bias term into the product $W^{(i)} X^{(i)}$ by replacing the the weight matrix with its augmentation by $B^{(i)}$ and $X^{(i)}$ by its augmentation by one. In order to make sure the solution can be inverted to the preceding layer by the activation function $A^{(i-1)}$ from that layer, we consider the following reparameterisation and define: 
\begin{equation}\label{opobjective}
    f(X) := || U ^{(i)} A^{(i-1)}(X) - V^{(i)} ||^2.
\end{equation}
By minimising this function with respect to $X$, we can obtain an approximate solution to both constraints \eqref{weightconstraint} and \eqref{gradconstraint} which is guaranteed to be invertible by $A^{(i-1)}$. If the layer in question is the first one in the network, then $f(X)$ will be defined as the following quadratic function without the reparameterisation term:
\begin{equation}\label{opobjectiveReduced}
    f(X) := || U ^{(i)}X - V^{(i)} ||^2.
\end{equation}

\paragraph{Pull-back constraint} In addition to the constraints defined by \eqref{weightconstraint} and \eqref{gradconstraint}, the weight constraint from the preceding layer might also impose constraint on the reconstruction in the current layer. This happens when the weight of the preceding layer has full-rank but the number of rows is greater than the number of columns. This will reduce the degree of freedom for the input to the current layer via the activation function. We can formulate this non-linear constraint mathematically and refer to it as the \textit{pull-back constraint}. The details of the formulation can be found in Appendix~\ref{AppendPullBackConstr}.

\paragraph{Optimisation problem in the convolutional case}
Based on the above considerations, the reconstruction problem for the convolutional layer can be formulated as solving the following optimisation problem:
\begin{equation}\label{convlayerobj}
    X^* =  \arg \min \left [ f(X) + ||g^{(i)} A^{(i)}(X)||^2 \right ],
\end{equation}
where $g^{(i)}$ is the pull-back constraint. When the layer in question is the first one in the network, then the function to be optimised in \eqref{convlayerobj} will reduce to $f(X)$ given in \eqref{opobjectiveReduced}.

\paragraph{COPA} We present the Combined Optimisation Attack (COPA) algorithm. It starts from the bottom of the network and iteratively solves for the input of a layer following \eqref{fullyConnectedGeneral} and \eqref{convlayerobj}. 
\begin{algorithm}[H]\label{COPA}
	\SetAlgoLined
	\KwIn{Label $Y_0$ for the target image; Gradients $\nabla_{W^{(i)}}J, 1 \leq i \leq d$ from the backward pass using the target image $X_0$;
    Initialise $W^{(i)}$ randomly for each $i$. Set $X^{(d)} = Y_0$.}
	\For{$i$ \textbf{in} $(d-1,...,0)$}{
	    Compute $\nabla_{X^{(i+1)}}J$ using \eqref{gradrelationsX}\;
	    Compute $\nabla_{Z^{(i)}}J$ from $\nabla_{X^{(i+1)}}J$ using \eqref{gradrelationsZ}\;
	    \textbf{if} the current layer is fully connected, solve for $X^{(i)}$ using \eqref{fullyConnectedGeneral}\;
	    \textbf{else if} the current layer is convolutional: 
	    
	    \ \ \ \ Define $U^{(i)}, V^{(i)}$ following \eqref{DefUandV}\;
	            
	    \ \ \ \ \textbf{if} $i = 0$, solve for $X^{(i)}$ using \eqref{opobjectiveReduced}\;
	    
	    \ \ \ \ \textbf{else if} $\rank(W^{(i-1)}) < |X^{(i)}|$, solve for $X^{(i)}$ using \eqref{opobjective}\;
	    
	    \ \ \ \ \textbf{else} solve for $X^{(i)}$ using \eqref{convlayerobj} \;
	}
	\KwResult{Input $X^{(0)}$ to the network.} 
	
	\caption{Constraint Optimisation Attack (COPA).}
\end{algorithm}

\paragraph{A security measure}
For a fully connected layer, we have shown in Lemma \ref{lemma} that we can always reconstruct the input in full. This can be regarded as no level of security and we omit it from our definition of the metric. For a convolutional layer, we consider $\rank(U) - |X|$ as an index to measure the efficacy of COPA. The larger this number is, the less rank-deficient the linear system \eqref{opobjective} is and so more likely for COPA to have a full reconstruction for this layer. We also notice that the position where the rank-deficiency happens also matters. The closer it is to the first layer, the bigger impact it has on the reconstruction. This is consistent with our intuition that if the representation of the input data loses information at the first layer, it will be unlikely to compensate for that loss in later layers. To accommodate for this effect, we discount the index $\rank(U) - |X|$ by the position of the layer in the network. Here is the definition of the proposed metric:
\begin{definition}
Suppose the model $M$ has $d$ convolutional layers indexed by $1, ... , d$, followed by a fully-connected layer. We define the following metric:
\begin{equation}
    c(M) : = \sum_{i = 1} ^d \frac{d-(i-1)}{d} \cdot \big( \rank(U^{(i)}) - n_i \big),
\end{equation}
where $U^{(i)}$ is defined in \eqref{DefUandV} and $n_i$ is the dimension of the input for the $i$-th layer.
\end{definition}
Because $\rank(U^{(i)}) \leq n_i$ for each convolutional layer, $c(M)$ will be non-positive. The larger the value of the metric is, the less secure the model tends to be and the more likely for COPA to create better reconstructions. The metric is better interpreted as an estimate of the security of the model against COPA. Our experiments have shown that it is possible to fully reconstruct the input using COPA or R-GAP when $c(M) = 0$.

\paragraph{Recommendations on architectural design}
Based on the proposed metric and analysis, we can see that a network tends to be less secure against COPA if it has wider convolutional layers that greatly increase the dimensions of the feature spaces at the beginning of the network and only shrinks the feature spaces towards the final layer. Overall, we would recommend designing a model with a small value of $c(M)$ if defendability against COPA-type attack is the main concern. 

\section{Experiments}\label{experiments}
We demonstrate the performance of COPA and how our proposed index $c(M)$ can be an indicator of model security in practice. We consider a three-layer network CNN3 consisting of two convolutional layers and a fully connected layer, and performing classification on CIFAR-10 \cite{krizhevsky2009learning} \cite{cifar10license}. For simplicity, we assume the convolutional layer to be bias-free and the fully-connected layer to have non-zero bias. However, these assumptions on bias are not necessary for the COPA to work. We have considered CNN3 because we need at least two convolutional layers to demonstrate the effect of the pull-back constraints. In order to illustrate typical architecture designs for the network, we consider four variations of the network, each representing a different case of changes in dimensions of intermediate feature spaces. These variants covers all possibilities of non-trivial changes of feature spaces in a three-layer CNN. For each network, weights are initialised randomly from a uniform distribution. The target image will go through one forward and backward pass to generate the gradients. As a post-processing step, we apply Total Variation denoising implemented by the scikit-image library \cite{scikit-image} with the weight set to 0.15. More results where the activations are chosen differently can be found in Appendix~\ref{AppendixFurtherResults}.
\begin{table}[h]
		\centering
		\scalebox{0.85}{
		\begin{tabular}{lcccc}
		\toprule
		     & Layer 1 & Layer 2 & Fully Connected & $c(M)$  \\
		     \midrule
		     CNN3 Variant 1 & 3,6,1 & 4,3,2 & \phantom{0}588 & -2267 \\
		     CNN3 Variant 2 & 4,6,2 & 3,3,2 & \phantom{0}147 & -1995 \\
		     CNN3 Variant 3 & 3,6,1 & 3,9,1 & 7056 & \phantom{-000}0 \\
		     CNN3 Variant 4 & 3,1,1 & 3,6,1 & 4704 & -2146 \\
		     \bottomrule
		\end{tabular}}
		\vspace{2mm}
		\caption{Model architecture for all four variants of the CNN3 model, with their $c(M)$. The numbers in columns `layer 1' and `layer 2' refer to kernel width, channels, strides accordingly. All of the variants have no padding, because padding does not play a decisive role in the efficacy of COPA. The numbers in the column `fully connected' refer to the input dimension of that fully-connected layer, whereas the output dimension is always 10.}\label{modelArch}
\vspace{-7mm}
\end{table}
\begin{figure}[h]
		\centering
		%
		%
		\begin{subfigure}{0.245\textwidth}
			\includegraphics[width=\linewidth]{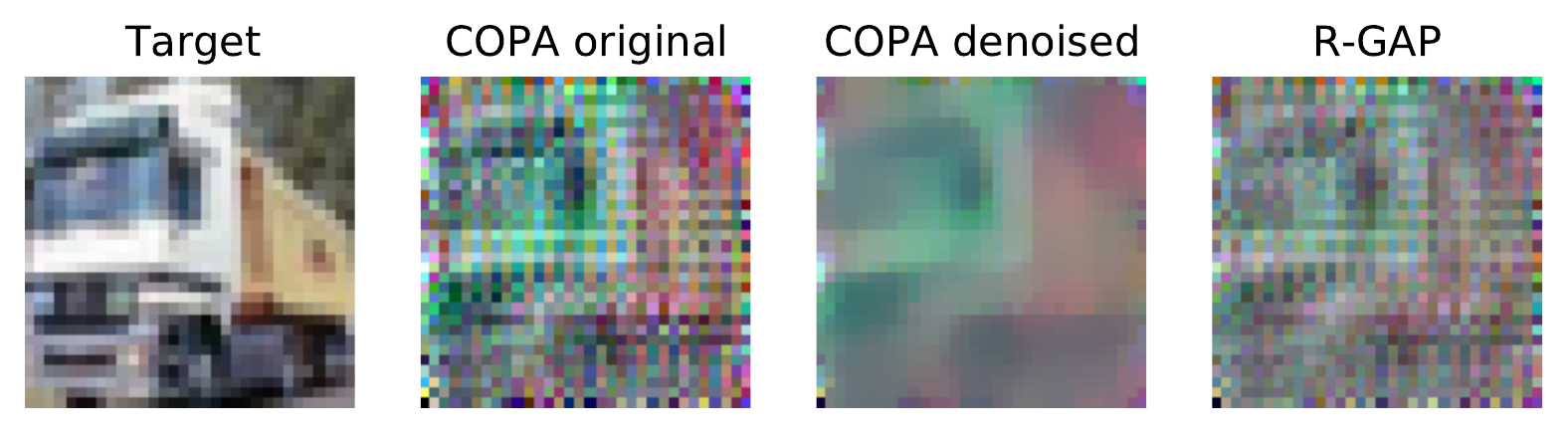}\\
			\includegraphics[width=\linewidth]{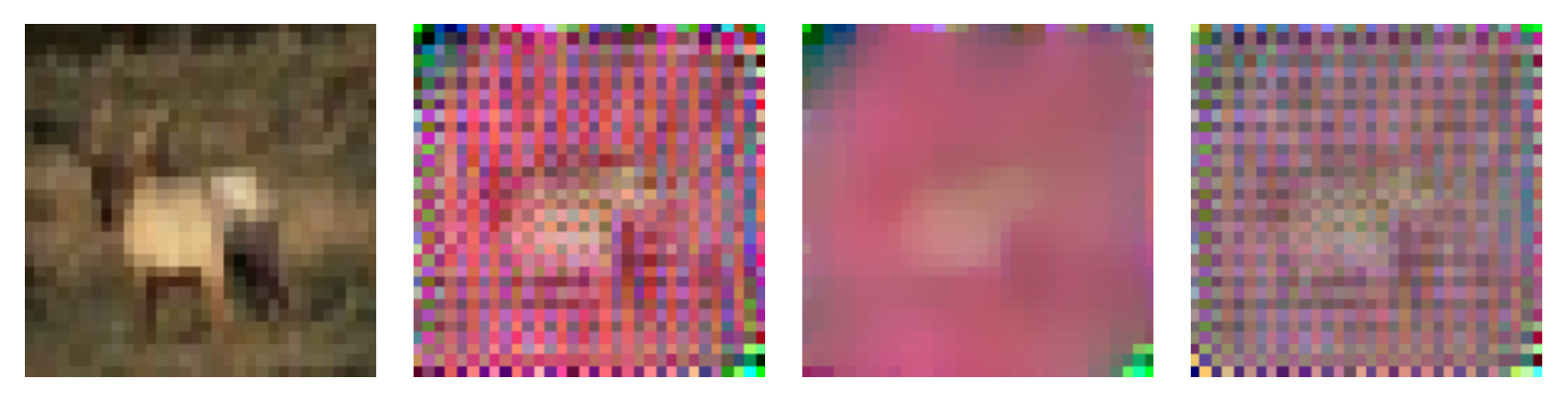}
			\caption{Variant 1}\label{fig:variant1Act1}
		\end{subfigure} 
		\begin{subfigure}{0.245\textwidth}
			\includegraphics[width=\linewidth]{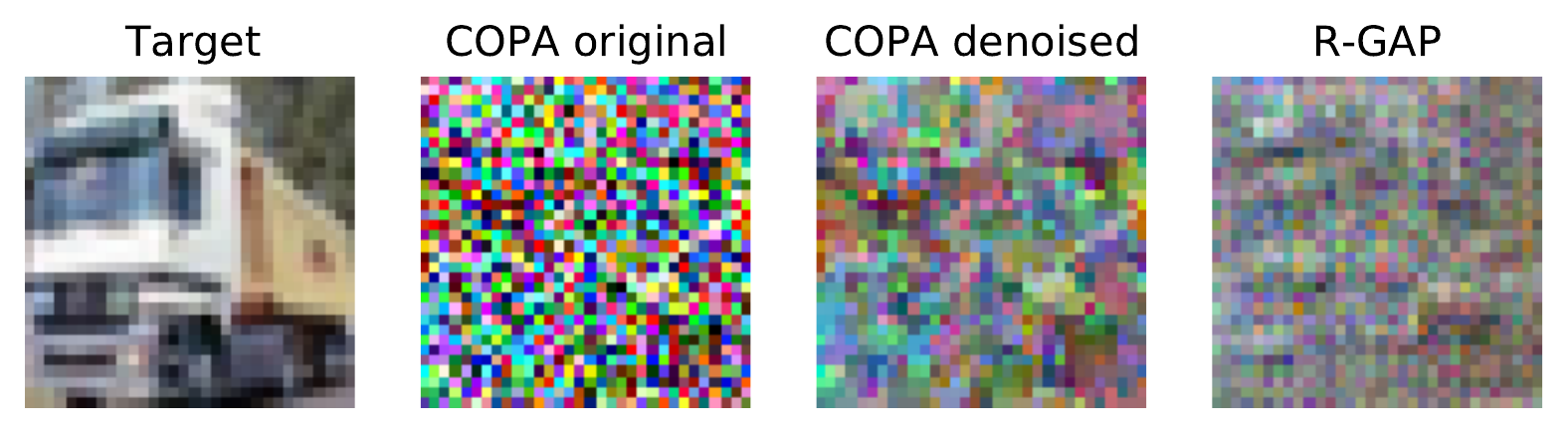}\\
			\includegraphics[width=\linewidth]{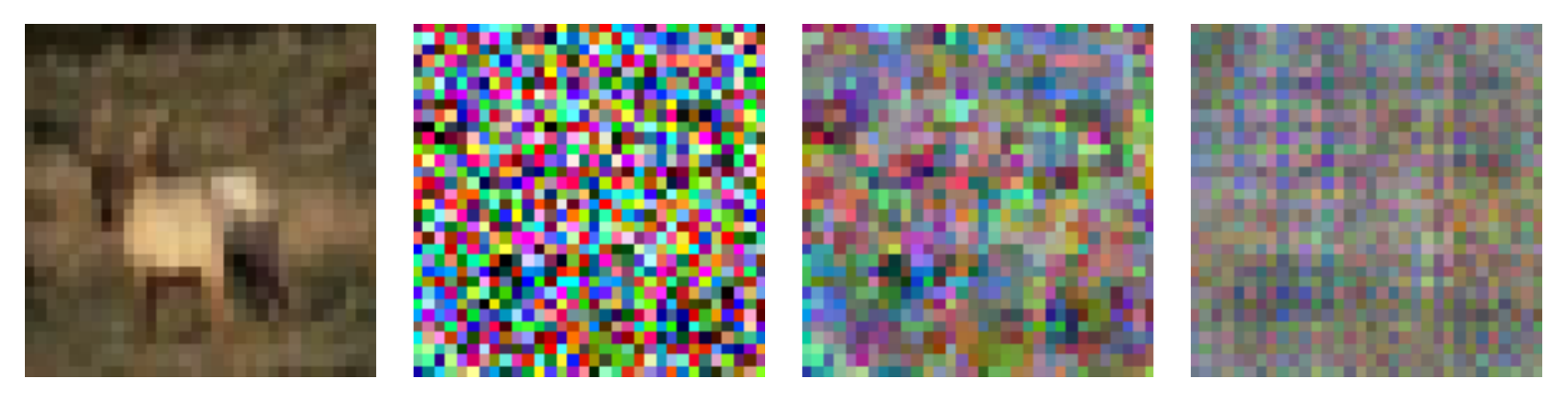}
			\caption{Variant 2}\label{fig:variantAct1}
		\end{subfigure} 
		\begin{subfigure}{0.245\textwidth}
			\includegraphics[width=\linewidth]{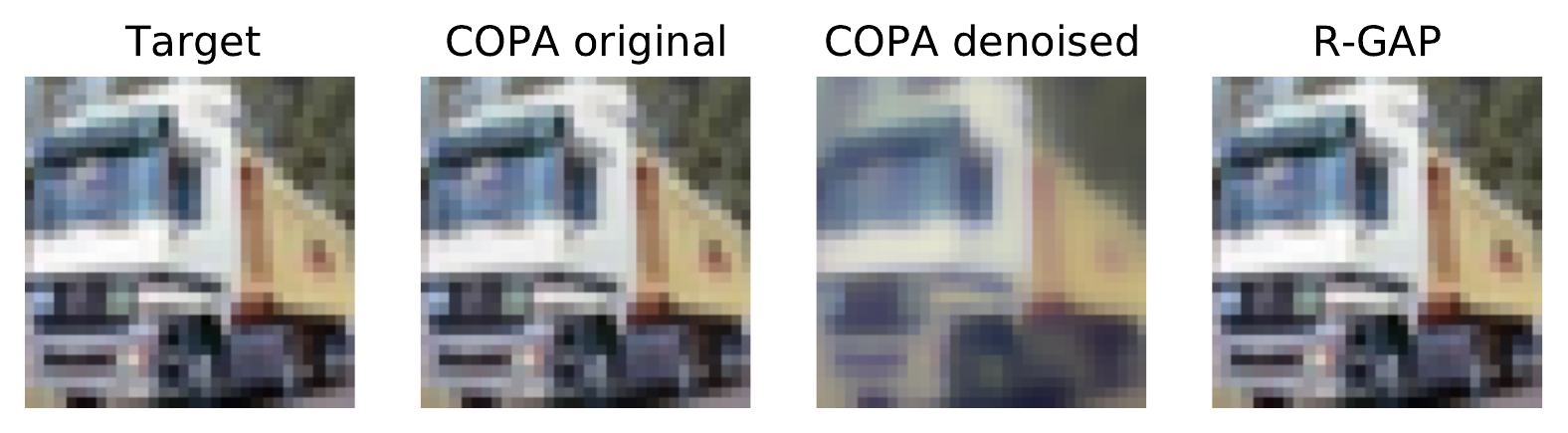}\\
			\includegraphics[width=\linewidth]{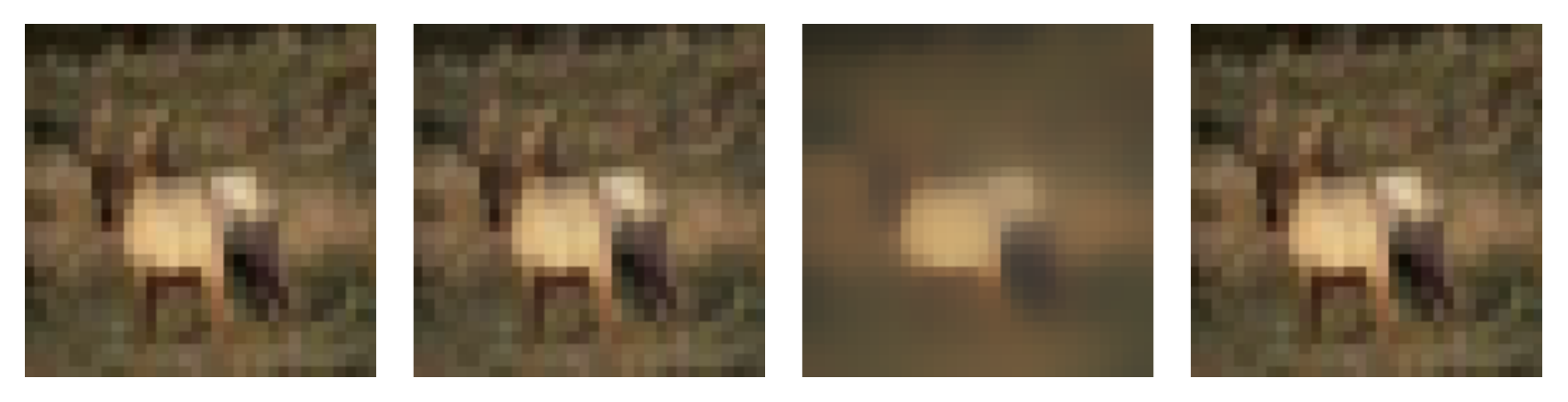}
			\caption{Variant 3}\label{fig:variant3Act1}
		\end{subfigure} 
		\begin{subfigure}{0.245\textwidth}
			\includegraphics[width=\linewidth]{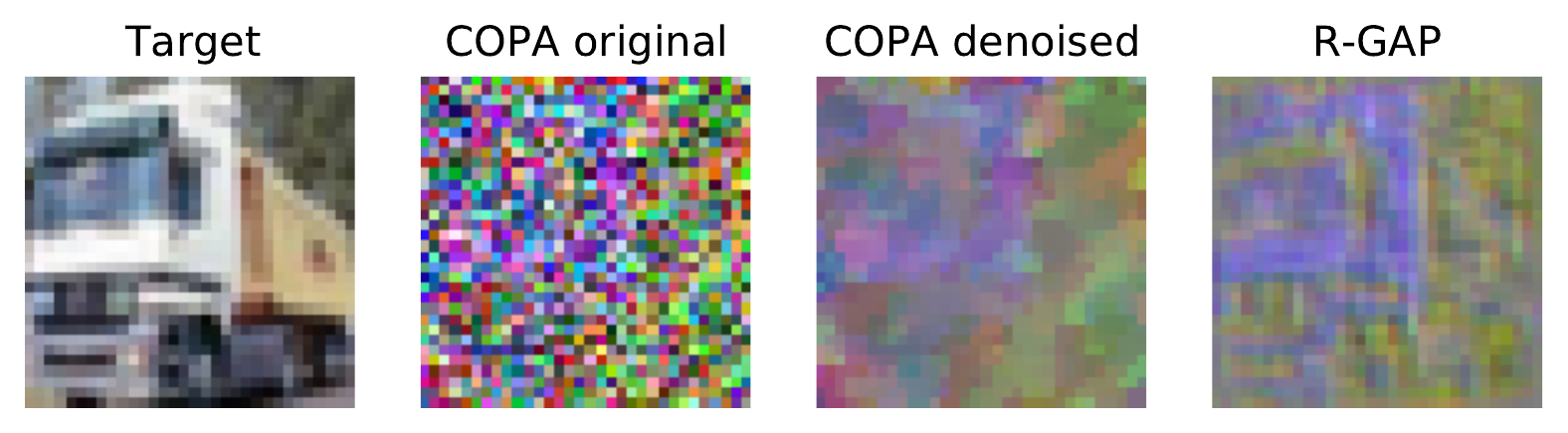}\\
			\includegraphics[width=\linewidth]{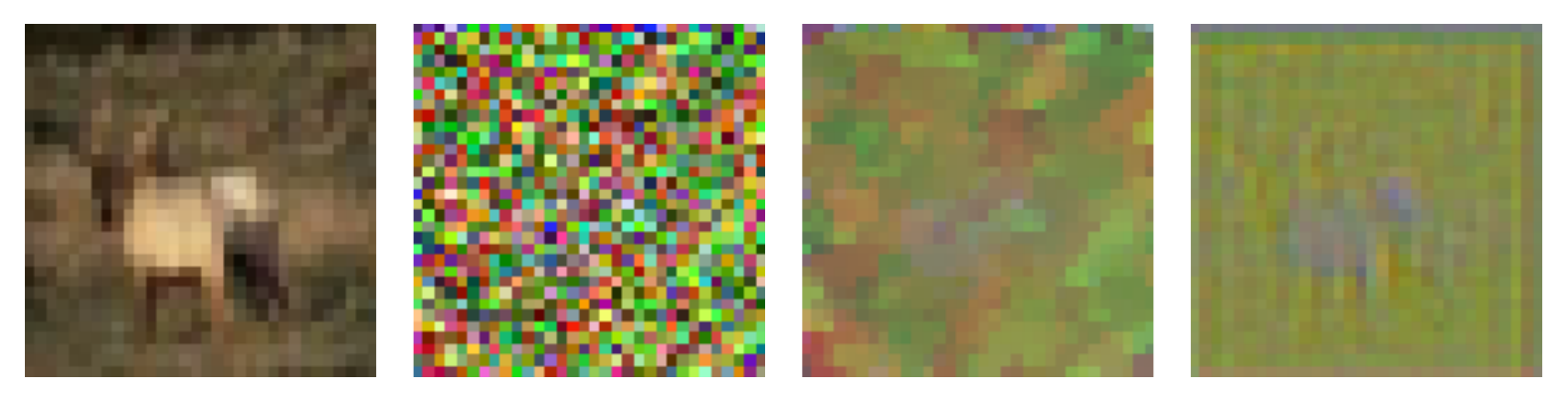}
			\caption{Variant 4}\label{fig:variant4Act1}
		\end{subfigure}
		\caption{Sample outputs from COPA applied to all four variants, when the activations are Tanh for all layers. Variant 1 and 3 have better reconstructions overall compared to Variant 2 and 4. This has been captured by our metric $c(M)$: the first layer of Variant 1 and 3 have contribution 0 (i.e. $\rank(U) = |X|$) to the metric; on top of that, Variant 3 has 0 contribution from the second layer as well, so it has the best reconstruction quality. There is little difference between COPA and R-GAP between Variant 1 and 3, showing limited impact from the pull-back constraint.}
		\label{fig:alltanh}
\vspace{-5mm}
\end{figure}
\section{Conclusion and future work}
In this paper, we advance our understanding of training-data leakage from gradients by developing a novel method COPA to reconstruct training data given gradients and the architecture of the target model. The key insight from COPA is formulating the reconstruction problem as solving linear systems at each layer formed by weights and gradients. This allows us to attribute the source of training-data leakage to the architecture of the model and we propose a metric to capture the level of security of the model against COPA. The metric can provide us a guideline in designing a deep network more securely. For future work, we will extend COPA to tackle batch-image reconstructions, against ResNet architecture, and against pre-trained models.

\section*{References}
\printbibliography[heading=none]

\newpage
\appendix
\begin{center}
\section*{Understanding Training-Data Leakage from Gradients in Neural Networks for Image Classification: Supplementary Material}
\end{center}
\section{Statement of solution and proof for the fully-connected layer}\label{appendReconFullyConnectedLayer}
\begin{lemma}\label{lemma}
If a fully connected layer has non-zero bias $B^{(i)} = (b_1 ^{(i)},...,b_m ^{(i)}) \in \BR^m$, then the input $X^{(i)} = (x_1 ^{(i)},...,x_n ^{(i)}) \in \BR^n$ is uniquely determined from the gradient constraint \eqref{gradconstraint}. Suppose $\exists k, 1 \leq k \leq m$, such that $b_k ^{(i)}\neq 0$ and $\frac{\partial J}{\partial b_k ^{(i)}} \neq 0$. Then $X^{(i)}$ is given by:
\begin{equation}
            x_l ^{(i)}= \frac{\partial J}{\partial w_{kl} ^{(i)} } \left (\frac{\partial J}{\partial b_k ^{(i)} } \right )^{-1}, \ \ 1 \leq l \leq n.
\end{equation}
More generally, assuming both $\frac{\partial J}{\partial x_k ^{(i+1)}}$ and $\frac{\partial a_k ^{(i)}}{\partial z_k ^{(i)}}$ are nonzero, we have 
\begin{equation}\label{fullyConnectedGeneral}
    x_l ^{(i)} = \frac{\partial J}{\partial w_{kl} ^{(i)}} \left (\frac{\partial J}{\partial x_k ^{(i+1)}} \right )^{-1} \left (\frac{\partial a_k ^{(i)}}{\partial z_k ^{(i)}} \right )^{-1}.
\end{equation}
\end{lemma}
The case of non-zero bias is due to \cite{aono2017privacy} and we show that it can be extended to the general case. 
\label{app:proof}
\begin{proof}
We will prove the general case first. By the construction of a fully-connected layer, we can take the $j$-th column of the gradient constraint \eqref{gradconstraint} which gives:
\begin{equation}
    (\frac{\partial J}{\partial w_{1j} ^{(i)}},...,\frac{\partial J}{\partial w_{nj} ^{(i)}})^T = x_j ^{(i)} (\frac{\partial J}{\partial z_1 ^{(i)}}, ... , \frac{\partial J}{\partial z_n ^{(i)}})^T.
\end{equation}
This implies that if $\frac{\partial J}{\partial z_k ^{(i)}} \neq 0$ for some $k, 1 \leq k \leq n$, then $X^{(i)}$ can be uniquely determined:
\begin{equation}
        x_j ^{(i)}= \frac{\partial J}{\partial w_{kj} ^{(i)}} (\frac{\partial J}{\partial z_k ^{(i)}})^{-1}.
\end{equation}
In the special case when $b_k ^{(i)}\neq 0$, we can see from the weight constraint \eqref{weightconstraint} that:
\begin{equation}
    \frac{\partial J}{\partial z_k ^{(i)}} = \frac{\partial J}{\partial b_k ^{(i)}},
\end{equation}
which was observed in \cite{aono2017privacy} and subsequently also in \cite{Fan2020}. In general, since the activation functions are assumed to be piecewise invertible and piecewise differentiable, and since $\frac{\partial J}{\partial x_k ^{(i+1)}}$ is assumed to be nonzero, we can compute $x_j ^{(i)}$ using \eqref{gradrelations}, which gives \eqref{fullyConnectedGeneral}.
\end{proof}

\section{Formulation of the pull-back constraint}\label{AppendPullBackConstr}
Consider the weight constraint from the preceding layer:
\begin{equation}\label{weightconstraintprevlayer}
    W^{(i-1)} X^{(i-1)} = Z^{(i-1)}.
\end{equation}
Let $m,n$ denote the number of rows and columns for $W^{(i-1)}$. For the moment, we view both $X^{(i-1)}$ and $Z^{(i-1)}$ as unknown. If $n < m$ and $W^{(i-1)}$ has full-rank, then $ n = \rank(W^{(i-1)}) < m = |Z^{(i-1)}|$, which implies that the degree of freedom for $Z^{(i-1)}$ will be smaller than $m$. This reduction of degree of freedom will impose a prior on the input to the current layer through the activation $A^{(i-1)}$ because $X^{(i)} = A^{(i-1)}(Z^{(i-1)})$. In the following discussion, we will express this prior as a non-linear constraint on $X^{(i)}$ which we call the \textit{pull-back constraint} (via the activation function).

We proceed by applying Singular Value Decomposition to $W^{(i-1)}$:
\begin{equation}\label{svdforW}
    W^{(i-1)} = R \begin{bmatrix}
                   \Sigma_n \\
                   0 \\
                  \end{bmatrix}
                  S^T,
\end{equation}
where $R$ and $S^T$ are unitary with sizes $m$ and $n$ respectively, $\Sigma_n$ is diagonal of size $n$ and the zero block has size $m-n$ by $n$. Notice that we do not require the singular values to be ordered, as long as they are all contained in the first $n$ rows in the block matrix. 

We can now express the weight constraint \eqref{weightconstraintprevlayer} as a single equation for $X^{(i)}$. To do that, first we define the following projection operator:
\begin{equation}
P^{(i-1)} := \begin{bmatrix}
        \I_{n} & 0 \\
        \end{bmatrix},
\end{equation}
where $\I_{n}$ is the identity matrix of size $n$ and the zero block has size $n$ by $m$. Now we rewrite \eqref{weightconstraintprevlayer} w.r.t. $Z^{(i-1)}$ using \eqref{svdforW}:
\begin{equation}\label{deductPbConstraint}
    W^{(i-1)} (\Sigma_n S^T)^{-1} P^{(i-1)} R^{-1} Z^{(i-1)} - Z^{(i-1)} = 0. 
\end{equation}
This can be further simplified into:
\begin{equation}
    R \begin{bmatrix}
        0 &  \\
          & \I_{m-n} \\
      \end{bmatrix} R^T Z^{(i-1)} = 0,
\end{equation}
where the block matrix has size $m$ by $m$ and we have used $R^T = R^{-1}$ since $R$ is unitary. 
We define an operator $g^{(i)}$ as follows:
\begin{equation}
    g^{(i)} := R \begin{bmatrix}
                0 &  \\
                  & \I_{m-n} \\
                \end{bmatrix} R^T (A^{(i)})^{-1}, 
\end{equation}
which will be non-linear if $A^{(i)}$ is non-linear. With this definition, the weight constraint \eqref{weightconstraintprevlayer} can now be written as a constraint on $X^{(i)}$: 
\begin{equation}\label{AddtionalConstraintPrevLayinG}
    g^{(i)}(X^{(i)}) = 0.
\end{equation}

In two special cases, the pull-back constraint will not be defined. The first case is when the layer in question is the first one in the network. The second one is when the weight $W^{(i-1)}$ is invertible or satisfies $\rank(W^{(i-1)}) = m < n$. 
\begin{remark}
If $W^{(i-1)}$ is rank-deficient and $\rank(W^{(i-1)}) < m$, there will still be a prior for $X^{(i)}$ for the same reason. However, in this case we cannot obtain the constraint as a function of $X^{(i)}$ because it also depends on $X^{(i-1)}$ which cannot be eliminated in the deduction of \eqref{deductPbConstraint}.
\end{remark}

\section{Further experimental results}\label{AppendixFurtherResults}
Here we provide results when the activation functions are LeakyReLU for layer 1 and 2, and Sigmoid for the last layer.
\begin{figure}[h]
		\centering
		%
		%
		\begin{subfigure}[h]{0.24\textwidth}
			\includegraphics[width=\linewidth]{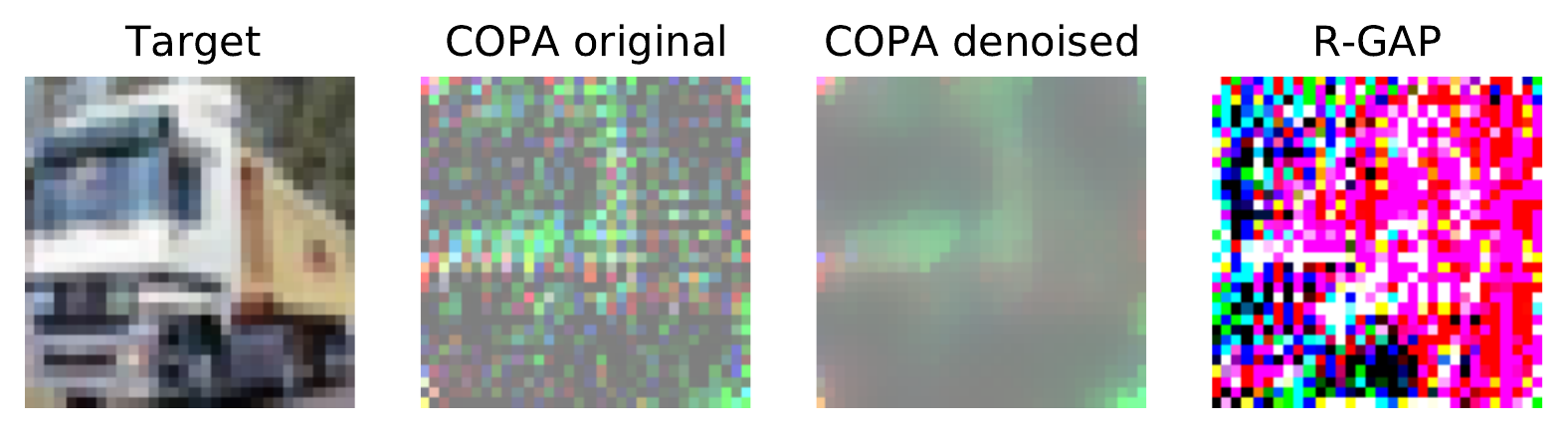}\\
			\includegraphics[width=\linewidth]{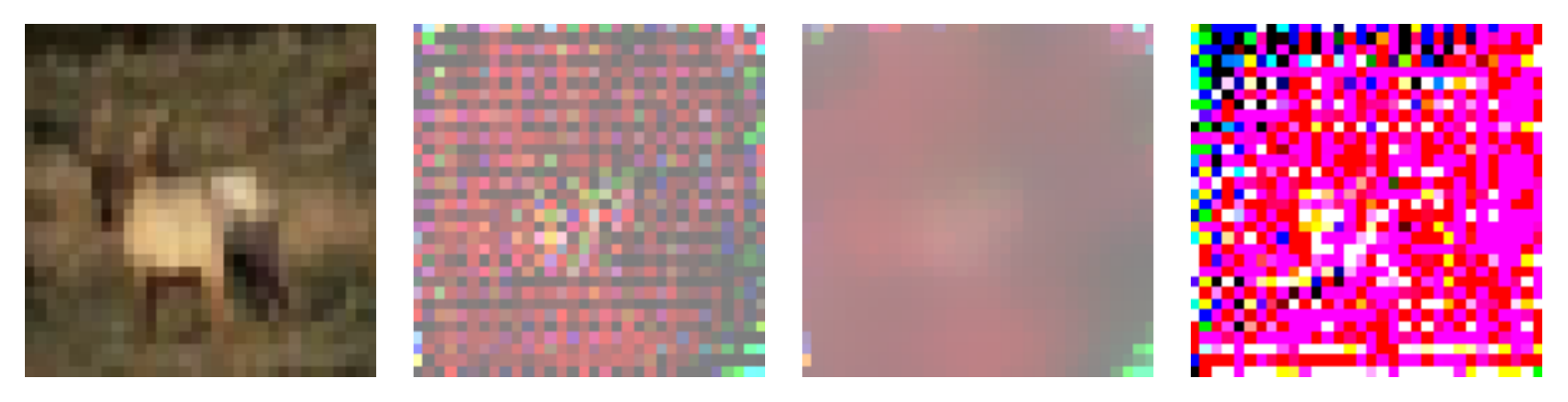}
			\caption{Variant 1}\label{fig:variant1Act2}
		\end{subfigure} 
		\begin{subfigure}[h]{0.24\textwidth}
			\includegraphics[width=\linewidth]{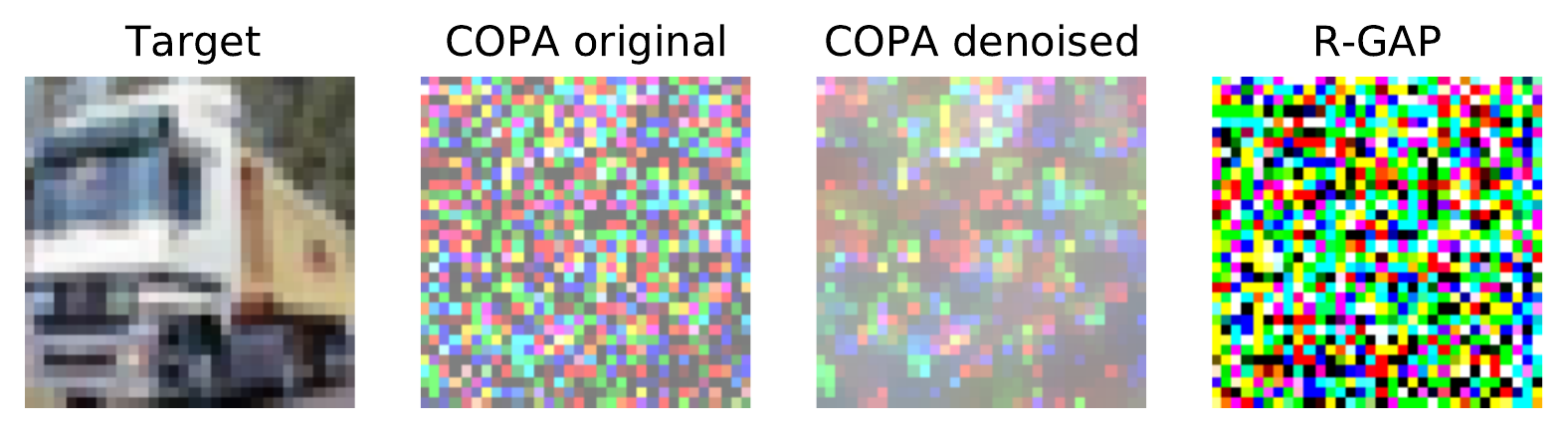}\\
			\includegraphics[width=\linewidth]{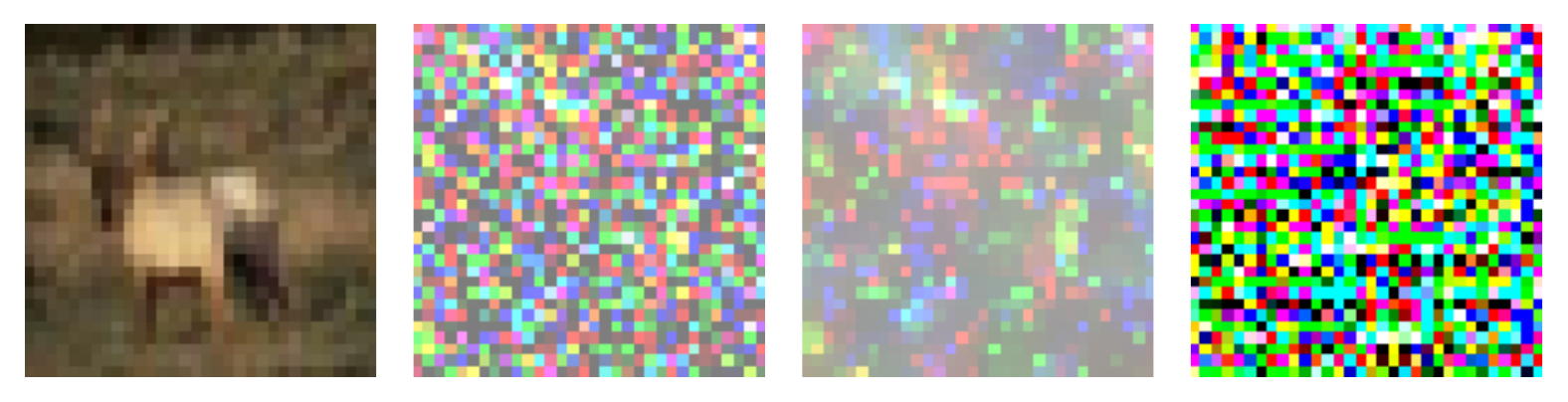}
			\caption{Variant 2}\label{fig:variant2Act2}
		\end{subfigure} \\
		\begin{subfigure}[h]{0.24\textwidth}
			\includegraphics[width=\linewidth]{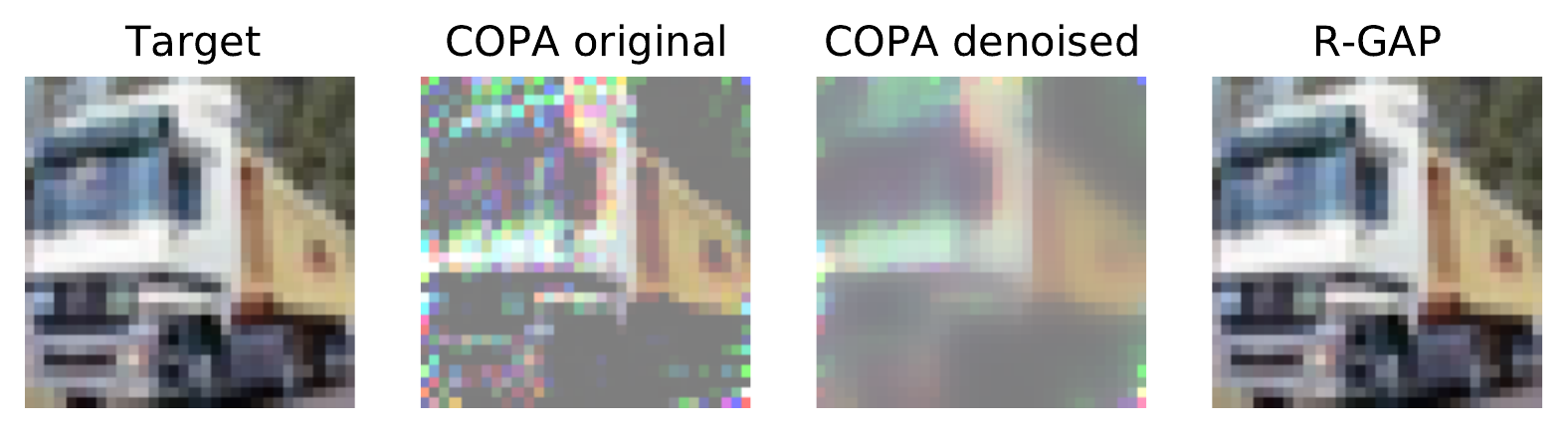}\\
			\includegraphics[width=\linewidth]{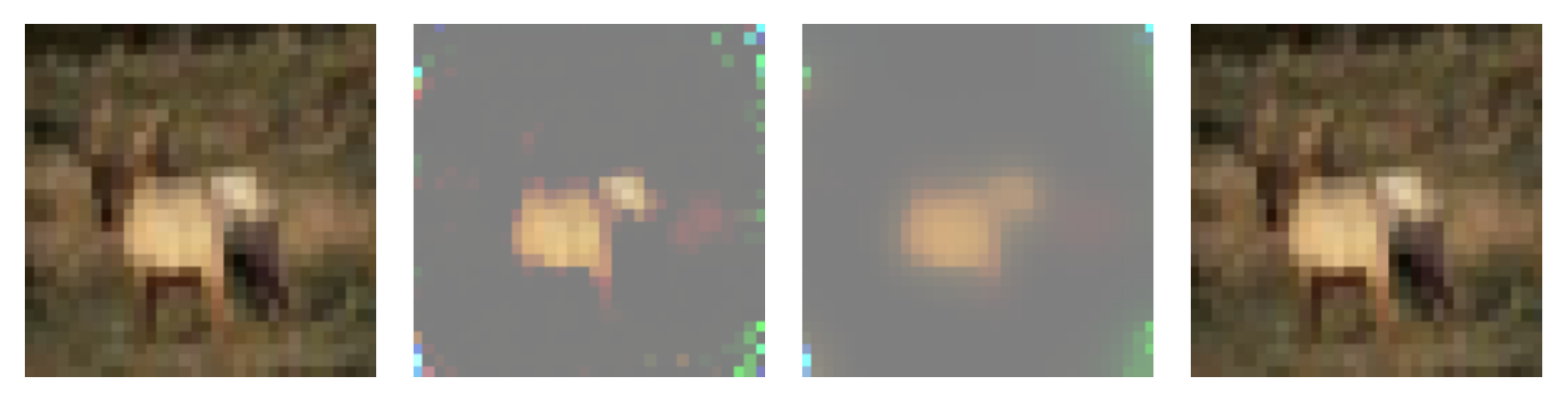}
			\caption{Variant 3}\label{fig:variant3Act2}
		\end{subfigure}
		\begin{subfigure}[h]{0.24\textwidth}
			\includegraphics[width=\linewidth]{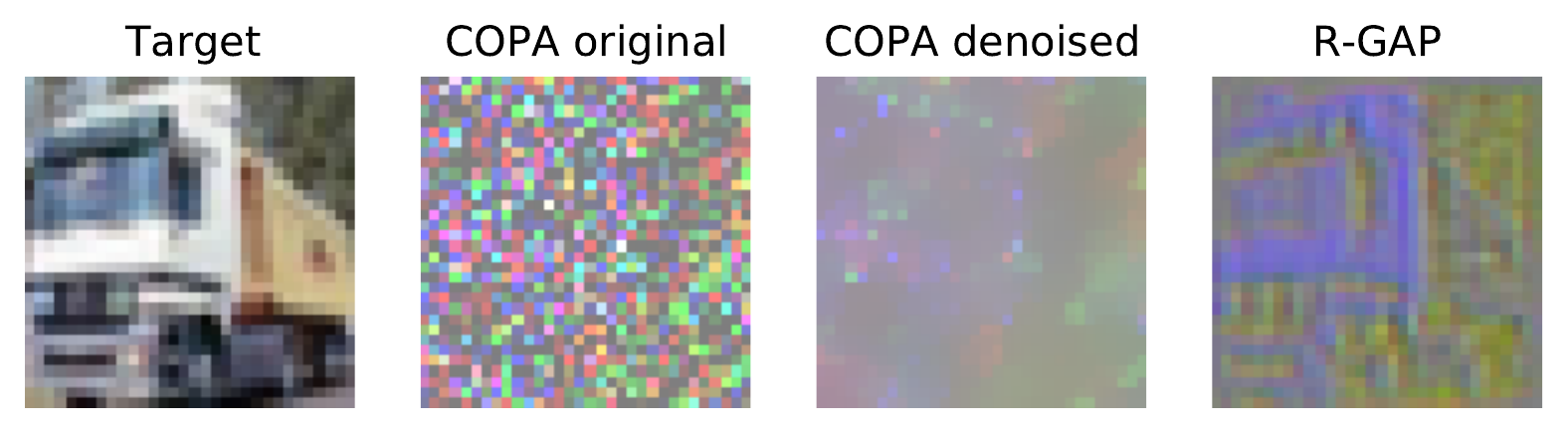}\\
			\includegraphics[width=\linewidth]{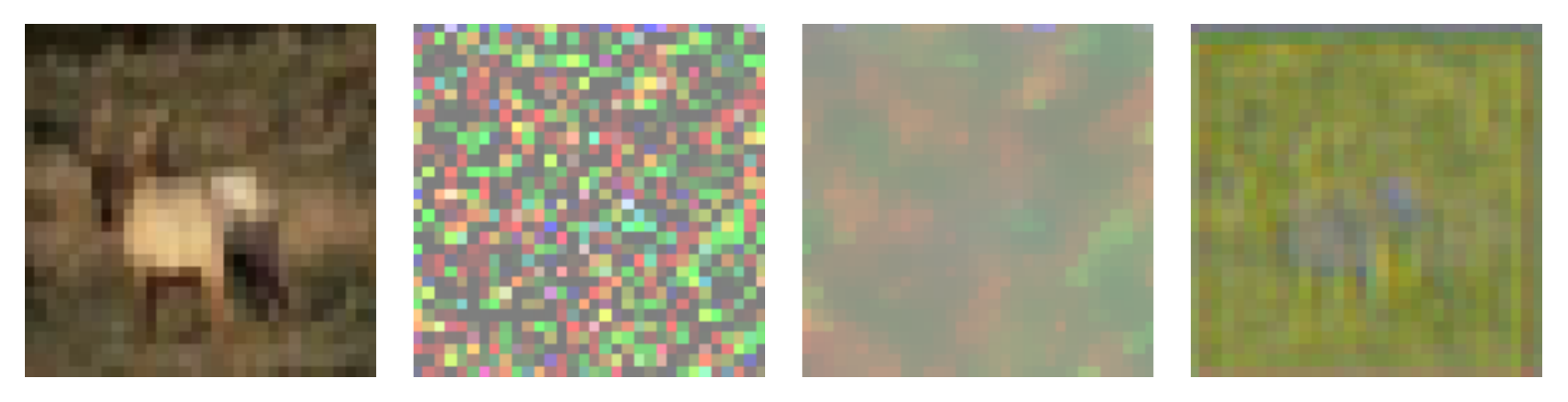}
			\caption{Variant 4}\label{fig:variant4Act2}
		\end{subfigure}
		\caption{Sample outputs from COPA applied to all four variants, when the activations are LeakyReLU for layer 1 and 2, and Sigmoid for the last layer. We can observe a consistent pattern compared to Figure \ref{fig:alltanh}.}
		\label{fig:lykysig}
\end{figure}
\newline
Furthermore, we have computed the MSE and PSNR for the outputs in Figure 1 and 2, to show that they are consistent with the visual observations (the numbers are MSE,PSNR respectively):
\begin{table}[ht]
		\centering
		\begin{tabular}[h]{lcccc}
		\toprule
		     Figure 1, variant 1 & COPA original & COPA denoised & R-GAP &  \\
		     \midrule
		     Truck & 0.0512, 61.0389 & 0.0426, 61.8361 & 0.0483, 61.2932 \\
		     Animal & 0.0766, 59.2867 & 0.0596, 60.3813 & 0.0578, 60.5147  \\
		     \bottomrule
		\end{tabular} 
		\begin{tabular}[h]{lcccc}
		\toprule
		     Figure 1, variant 2 & COPA original & COPA denoised & R-GAP &  \\
		     \midrule
		     Truck & 0.1921, 55.2966 & 0.0738, 59.4519 & 0.0502, 61.1218 \\
		     Animal & 0.1913, 55.3148 & 0.0707, 59.6379 & 0.0533, 60.8664  \\
		     \bottomrule
		\end{tabular}
		\begin{tabular}[h]{lcccc}
		\toprule
		     Figure 1, variant 3 & COPA original & COPA denoised & R-GAP &  \\
		     \midrule
		     Truck & 0.0001, 87.9242 & 0.0088, 68.6745 & 0.0000, 181.5003 \\
		     Animal & 0.0001, 90.5780 & 0.0031, 73.2162 & 0.0000, 180.6153  \\
		     \bottomrule
		\end{tabular}
		\begin{tabular}[h]{lcccc}
		\toprule
		     Figure 1, variant 4 & COPA original & COPA denoised & R-GAP &  \\
		     \midrule
		     Truck & 0.1007, 58.1021 & 0.0524, 60.9338 & 0.0472, 61.3900 \\
		     Animal & 0.0861, 58.7818 & 0.0373, 62.4189 & 0.0385, 62.2730  \\
		     \bottomrule
		\end{tabular}
		\begin{tabular}[h]{lcccc}
		\toprule
		     Figure 2, variant 1 & COPA original & COPA denoised & R-GAP &  \\
		     \midrule
		     Truck & 0.0546, 60.7559 & 0.0516, 61.0035 & 25.2923, 34.1009 \\
		     Animal & 0.1046, 57.9361 & 0.0950, 58.3548 & 31.3128, 33.1736  \\
		     \bottomrule
		\end{tabular}
		\begin{tabular}[h]{lcccc}
		\toprule
		     Figure 2, variant 2 & COPA original & COPA denoised & R-GAP &  \\
		     \midrule
		     Truck & 0.1751, 55.6978 & 0.1016, 58.0623 & 12.3527, 37.2132 \\
		     Animal & 0.2571, 54.0298 & 0.1822, 55.5247 & 3.6857, 42.4656  \\
		     \bottomrule
		\end{tabular}
		\begin{tabular}[h]{lcccc}
		\toprule
		     Figure 2, variant 3 & COPA original & COPA denoised & R-GAP &  \\
		     \midrule
		     Truck & 0.0355, 62.6284 & 0.0386, 62.2610 & 0.0000, 137.5824 \\
		     Animal & 0.0565, 60.6117 & 0.0563, 60.6291 & 0.0000, 143.4470  \\
		     \bottomrule
		\end{tabular}
		\begin{tabular}[h]{lcccc}
		\toprule
		     Figure 2, variant 4 & COPA original & COPA denoised & R-GAP &  \\
		     \midrule
		     Truck & 0.0868, 58.7466 & 0.0630, 60.1398 & 0.0483, 61.2918 \\
		     Animal & 0.1061, 57.8722 & 0.0908, 58.5509 & 0.0382, 62.3075  \\
		     \bottomrule
		\end{tabular}
		\vspace{2mm}
		\caption{MSE and PSNR for the outputs in Figure 1 and 2 for all variants. In each cell, the numbers are MSE and PSNR correspondingly.}
\end{table}
\end{document}